\documentclass[final]{cvpr}

\usepackage{times}
\usepackage{epsfig}
\usepackage{graphicx}
\usepackage{amsmath}
\usepackage{amssymb}
\usepackage{multirow}
\usepackage{booktabs}

\newcommand{\squeezeup}{\vspace{-2mm}}

\usepackage[pagebackref=true,breaklinks=true,colorlinks,bookmarks=false]{hyperref}

\def\cvprPaperID{1228} % *** Enter the CVPR Paper ID here
\def\confYear{CVPR 2021}

\begin{document}

%%%%%%%%% TITLE
\title{Dual-stream Multiple Instance Learning Network\\for Whole Slide Image Classification with Self-supervised Contrastive Learning}

\author{Bin Li\textsuperscript{1,2}, Yin Li\textsuperscript{3,4*}, Kevin W. Eliceiri\textsuperscript{1,2,5*}\\
\textsuperscript{1}Department of Biomedical Engineering, University of Wisconsin-Madison\\
\textsuperscript{2}Morgridge Institute for Research, Madison, WI USA\\
\textsuperscript{3}Department of Biostatistics and Medical Informatics, University of Wisconsin-Madison\\
\textsuperscript{4}Department of Computer Sciences, University of Wisconsin-Madison\\
\textsuperscript{5}Department of Medical Physics, University of Wisconsin-Madison\\
{\tt\small \{bli346, yin.li, eliceiri\}@wisc.edu}\\
%{\tt\small *co-corresponding authors}
% \and
% Yin Li\\
% University of Wisconsin at Madison\\
% Department of Biostatistics and Medical Informatics\\
% {\tt\small yin.li@wisc.edu}
% \and
% Kevin W. Eliceiri\\
% Morgridge Institute for Research\\
% Department of Medical Physics, University of Wisconsin-Madison\\
% {\tt\small eliceiri@wisc.edu}
}

\maketitle
{\let\thefootnote\relax\footnote{* Co-corresponding authors.}}
% Remove page numbers
\pagestyle{empty}
\thispagestyle{empty}

%%%%%%%%% ABSTRACT
\squeezeup
\begin{abstract}
\squeezeup
 We address the challenging problem of whole slide image (WSI) classification. 
 WSIs have very high resolutions and usually lack localized annotations. 
 WSI classification can be cast as a multiple instance learning (MIL) problem when only slide-level labels are available. 
 We propose a MIL-based method for WSI classification and tumor detection that does not require localized annotations. 
 Our method has three major components.
 First, we introduce a novel MIL aggregator that models the relations of the instances in a dual-stream architecture with trainable distance measurement. 
 Second, since WSIs can produce large or unbalanced bags that hinder the training of MIL models, we propose to use self-supervised contrastive learning to extract good representations for MIL and alleviate the issue of prohibitive memory cost for large bags. 
 Third, we adopt a pyramidal fusion mechanism for multiscale WSI features, and further improve the accuracy of classification and localization.
 Our model is evaluated on two representative WSI datasets.
 The classification accuracy of our model compares favorably to fully-supervised methods, with less than 2\% accuracy gap across datasets. 
 Our results also outperform all previous MIL-based methods. 
 Additional benchmark results on standard MIL datasets further demonstrate the superior performance of our MIL aggregator on general MIL problems. GitHub repository: \url{https://github.com/binli123/dsmil-wsi}
\end{abstract}

%%%%%%%%% BODY TEXT
\squeezeup
\section{Introduction}
\label{sec:intro}
% motivation: why WSI
Whole slide scanning is a powerful and widely used tool to visualize tissue sections in disease diagnosis, medical education, and pathological research~\cite{cornish_whole-slide_2012, pantanowitz_review_2011}. 
The scanning converts tissues on glass slides into digital whole slide images (WSIs) for assessment, sharing, and analysis. Automated disease detection in WSIs has been a long-standing challenge for computer aided diagnostic systems. We have begun to see some recent success from computer vision and medical image analysis communities~\cite{chen_augmented_2019, sirinukunwattana_gland_2017, campanella_clinical-grade_2019, keikhosravi_non-disruptive_2020, li_single_2020, litjens_survey_2017}, fueled by the advances in deep learning.

% background: patch-based WSI classification
WSIs have extremely high resolutions --- a typical pathology image has a size of $40,000 \times 40,000$. Consequently, the most widely used paradigm for WSI classification is patch-based processing --- a WSI is divided into thousands of small patches and further examined by a classifier \eg., a convolutional neural network (CNN)~\cite{hou_patch-based_2016, xu_deep_2015, mousavi_automated_2015, cruz-roa_automatic_2014, maksoud2020sos}. In clinics, a disease-positive tissue section might only take a small portion (\eg, less than 20\%) of the whole tissue, leading to a large number of disease-negative patches. Unfortunately, with gigapixel resolution, patch-level labeling by expert pathologists is very time consuming and difficult to scale. To address this challenge, several recent studies~\cite{hou_patch-based_2016, campanella_clinical-grade_2019, hashimoto_multi-scale_2020} have demonstrated the promise of weakly supervised WSI classification, where only slide-level labels are used to train a patch-based classifier.

% the teaser figure
\begin{figure}
    \centering
    \includegraphics[width=0.45\textwidth]{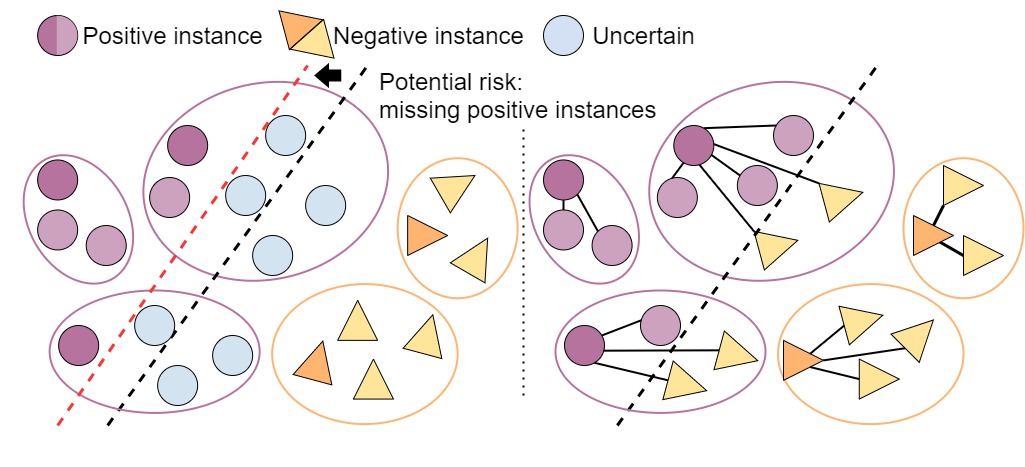}
    \caption{Decision boundary learned in MIL. \textbf{Left:} Max pooling delineates the decision boundary according to the highest-score instances in each bag. \textbf{Right:} DSMIL measures the distance between each instance and the highest-score instance.}
    \label{fig:slmil}\vspace{-1.5em}
\end{figure}

% context of multiple instance learning
The majority of previous approaches~\cite{hou_patch-based_2016, xu_deep_2015, mousavi_automated_2015, cruz-roa_automatic_2014, hashimoto_multi-scale_2020, chikontwe_multiple_2020} on weakly supervised WSI classification follows a multiple instance learning (MIL) problem formulation~\cite{dietterich_solving_1997, maron_framework_1998}, where each WSI is considered as a \emph{bag} that contains many \emph{instances} of patches. A WSI (bag) is labeled as disease-positive if any of its patches (instances) is disease-positive (\eg, with lesions). Patch-level features or scores are extracted, aggregated, and examined by a classifier that predicts slide-level labels. 
% Once trained, this patch-based classifier can be used to label the full image. 
Recent MIL based approaches have greatly benefited from using deep neural networks for feature extraction and feature aggregation~\cite{ilse_attention-based_2018, wang_deep_2016, oquab_is_2015}. 

% main challenges of current approaches
Two major challenges exist in developing deep MIL models for weakly supervised WSI classification. First, when patches (instances) in positive images (bags) are highly unbalanced, \ie, only a small portion of patches are positive, the models are likely to misclassify those positive instances~\cite{ilse_attention-based_2018} when using a simple aggregation operation, such as the widely adopted max-pooling. This is because, under the assumptions of MIL, max-pooling can lead to a shift of the decision boundary compared to fully-supervised training (Figure~\ref{fig:slmil}). Besides, the model can easily suffer from overfitting and unable to learn rich feature representations due to the weak supervisory signal~\cite{dehaene_self-supervision_2020, lu_semi-supervised_2019, akbar_cluster-based_2018}. 
Second, current models either use fixed patch features extracted by a CNN or only update the feature extractor using a few high score patches, as the end-to-end training of the feature extractor and aggregator is prohibitively expensive for large bags~\cite{dehaene_self-supervision_2020, campanella_clinical-grade_2019, lu_semi-supervised_2019}. Such a simplified learning scheme might lead to sub-optimal patch features for WSI classification.

% how we will solve the challenges -- our method
To address these challenges, we propose a novel deep MIL model, dubbed dual-stream multiple instance learning network (DSMIL). Specifically, DSMIL jointly learns a patch (instance) and an image (bag) classifier, using a two-stream architecture. The first stream deploys a standard max-pooling to identify the highest scored instance (referred to as \emph{critical instance}), while the second stream computes an attention score for each instance by measuring its distance to the critical instance. DSMIL further applies a soft selection of instances using the attention scores, leading to a decision boundary that better delineates the instances in positive bags, as shown in Figure~\ref{fig:slmil}. Importantly, DSMIL makes use of self-supervised contrastive learning for training the feature extractor for WSI, producing strong patch representations. In addition, DSMIL incorporates a multiscale feature fusion mechanism that can leverage tissue features ranging from millimeter-scale (\eg, vessels and glands) to cellular-scale (tissue microenvironment).

% hightlight the results
We evaluate DSMIL for weakly supervised WSI classification on two public WSI datasets including Camelyon16 and TCGA lung cancer. The results show that DSMIL outperforms other recent MIL models in classification accuracy by at least 2.3\%. More importantly, our classification accuracy compares favorably to fully-supervised methods, with less than 2\% accuracy gap. Moreover, DSMIL also has superior localization accuracy, outperforming previous MIL models by a significant margin. Finally, we demonstrate the state-of-the-art performance of DSMIL on general MIL problems beyond weakly supervised WSI classification.

\section{Related Work}
\label{sec:related_work}
Our work develops MIL for WSI analysis using deep models. MIL itself is a well-established topic. We refer the readers to~\cite{carbonneau2018multiple} for a survey. In this section, we briefly review recent efforts on deep MIL models, as well as relevant works on MIL models for WSI analysis. 

\noindent \textbf{Deep MIL Models}.
Conventionally, MIL models consider handcrafted aggregators, such as mean-pooling and max-pooling~\cite{feng_deep_2017, pinheiro_image-level_2015}. 
Recently, it is shown that parameterizing the aggregation operator with neural networks can still be beneficial~\cite{feng_deep_2017, wang_revisiting_2018, oquab_is_2015}. 
Ilse \etal~\cite{ilse_attention-based_2018} proposed an attention-based aggregation operator parameterized by neural networks which includes the contribution of each instance to the bag embedding. 
Methods that consider the contextual information are proposed to model the dependencies between the instances %and the instances to the bag, 
such as graph neural network-based approaches and capsule network-based approaches~\cite{tu_multiple_2019, yan_deep_2018, chikontwe_multiple_2020}. 

We deploy a non-local operation to model the instance-to-instance and instance-to-bag relations~\cite{wang_non-local_2018}. 
Differing from the attention mechanism in attention-based MIL (ABMIL)~\cite{ilse_attention-based_2018}, the attentions in our model are explicitly computed based on a trainable distance measurement. 
Our method is also different from graph models and capsule networks in that the weights between the nodes are functions of the two nodes instead of learned parameters~\cite{scarselli_graph_2009, sabour_dynamic_2017}. 
The measurement mechanism is similar to self-attention~\cite{vaswani_attention_2017, wang_non-local_2018}, but differs in that the measurement is done only between one node (the critical instance) to the others. Our dual-stream non-local operation also differs from asymmetric non-local operation in that the embeddings are filtered according to the confidence scores learned in a separate branch, instead of on the embeddings~\cite{zhu_asymmetric_2019}.
In addition, deep MIL models have been considered for other weakly supervised vision tasks, including weakly supervised object localization~\cite{cinbis2016weakly} and detection~\cite{tang2017multiple,wan2019c}. In this paper, we focus on weakly supervised classification of WSI.

\noindent \textbf{MIL Models for WSI Analysis}.
MIL has been successfully applied to WSI analysis for tasks such as cell segmentation and tumor detection~\cite{xu2014weakly, hou_patch-based_2016, quellec_multiple-instance_2017, kandemir_computer-aided_2015, campanella_clinical-grade_2019, chikontwe_multiple_2020}. 
Campanella \etal~\cite{campanella_clinical-grade_2019} show that a MIL classifier trained on large weakly-labeled WSI datasets generalizes better than a fully-supervised classifier trained on pixel-level-annotated small lab datasets. 
The former is easy to obtain on large scale from everyday clinics while the latter requires labor-intensive annotations in research labs. 

\begin{figure*}[t!]
    \centering
    \includegraphics[width=0.9\textwidth]{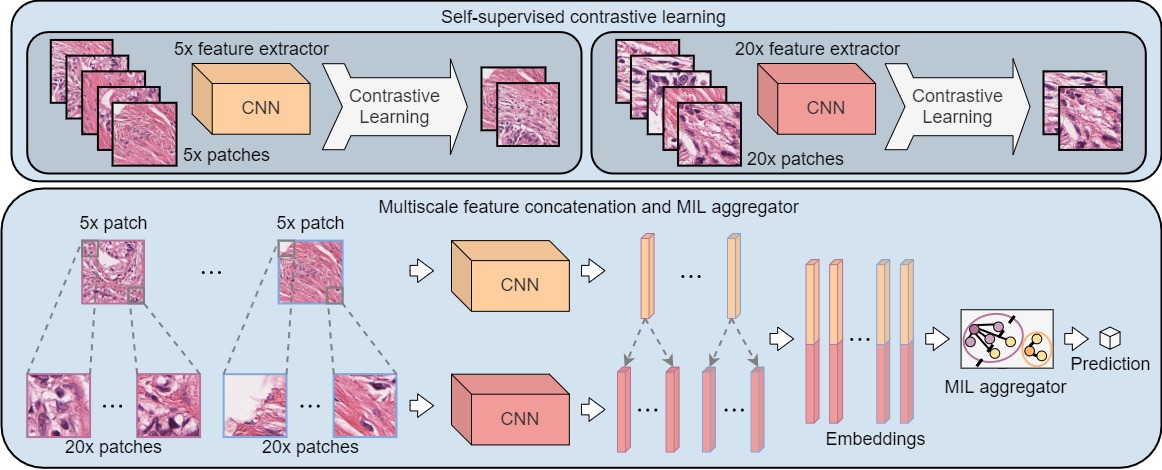}
    \caption{Overview of our DSMIL. Patches extracted from each magnification of the WSIs are used for self-supervised contrastive learning separately. The trained feature extractors are used to compute embeddings of patches. Embeddings of different scales of a WSI are concatenated to form feature pyramids to train the MIL aggregator. The figure shows an example of two magnifications (20$\times$ and 5$\times$). The 5$\times$ feature vector is duplicated and concatenated with each of the 20$\times$ feature vectors of the sub-images within this 5$\times$ patch.}
    \label{fig:dsmil}\vspace{-1.5em}
\end{figure*}

Training a CNN for good feature representations in MIL is non-trivial for WSI analysis, due to the prohibitive memory requirement and the noisy supervisory signal~\cite{lu_semi-supervised_2019, dehaene_self-supervision_2020}. Recently, semi-supervised learning has been used to enable the training of the classifier for WSI classification with limited patch-level labels \cite{koohbanani_self-path_2020}. In contrast, our work makes use of self-supervised contrastive learning~\cite{chen_simple_2020} for feature extraction in MIL. Self-supervised contrastive learning has demonstrated success in learning visual representations~\cite{oord2018representation, chen_simple_2020,he2020momentum}, yet remains unexplored in WSI analysis.

The assessment of WSIs by pathologists is done in multiscale~\cite{bejnordi_multi-scale_2015, gao_multi-scale_2016, tokunaga_adaptive_2019} and it is common to consider multiscale features in WSI analysis. Using bags that simply include features from different magnifications of WSI in MIL has shown to be beneficial~\cite{hashimoto_multi-scale_2020}. Another possibility~\cite{maksoud2020sos} is to select regions at low-magnification and further zoom in these regions for high-magnification patches. Our multiscale feature analysis strategy is inspired by previous works on multiscale feature representation using deep models~\cite{ronneberger2015u,lin2017feature}, yet simultaneously benefits our DSMIL model for the ability to locally-constrain the patch attentions.

% \yin{Need to discuss / mention \cite{maksoud2020sos} in this section.}

\section{Method}
\label{sec:method}
We now present our method for weakly supervised WSI classification. This section introduces the formulation of MIL and presents our model --- DSMIL.

% \yin{Sec 3.2 - 3.4 need to be re-organized. DSMIL is the full model, including the dual stream MIL, contrastive learning and multi-scale feature representation. The current writing reads as if they are totally separate. I'd suggest adding a short overview of the model at the start of 3.2 that sticks all components together, merging Sec 3.2-3.4, and adding overview figure of the full model. You can modify the current Fig 3 for the overview figure. Fig 2 is all about tech details and are not very interesting. Consider using a single column (smaller) figure here. You should probably highlight the new overview figure rather than the current figure 2.}

\subsection{Background: MIL Formulation}
In MIL, a group of training samples is considered as a bag containing multiple instances. 
Each bag has a bag label that is positive if the bag contains at least one positive instance and negative if it contains no such positive instance. The instance-level labels are unknown.
In the case of binary classification, let $B=\{(x_1, y_1),...,(x_n,y_n)\}$ be a bag where $x_i \in \chi$ are instances with labels $y_i \in \{0, 1\}$, the label of $B$ is given by 
\begin{equation}
\small
    c(B)=    
    \begin{cases}
        0, &\text{iff $\sum{y_i}=0$}\\
        1, &\text{otherwise}
    \end{cases}
\label{eq:bag_1}
\end{equation}
MIL further uses a suitable transformation $f$ and a \emph{permutation-invariant} transformation $g$ \cite{ilse_attention-based_2018, charles_pointnet_2017} to predict the label of $B$, given by
% explain permutation-invariant somewhere in the introduction
\begin{equation} \label{eq:mil}
\small
    c(B)=g(f(x_0),...,f(x_n))
\end{equation}

Multiple instance learning could be modeled in two ways based on the choices of $f$ and $g$: 1) Instance-based approach. $f$ is an instance classifier that scores each instance, $g$ is a pooling operator that aggregates the instance scores to produce a bag score. 2) Embedding-based approach. $f$ is an instance-level feature extractor that maps each instance to an embedding, $g$ is an aggregation operator that produces a bag embedding from the instance embeddings and outputs a bag score based on the bag embedding.
% It is shown that the embedding-based approach generally performs better in terms of the bag level classification accuracy \cite{wang_revisiting_2018}. 
The embedding-based method produces a bag score based on a bag embedding directly supervised by the bag label and usually yields better accuracy compared to the instance-based method~\cite{wang_revisiting_2018}, however, it is usually harder to determine the key instances that trigger the classifier~\cite{liu_detecting_2017}. 

In the setting of weakly supervised WSI classification, each WSI is considered as a bag and the patches extracted from it are considered as the instances of this bag. We will then describe our model that jointly learns a instance-level classifier as well as an embedding aggregator and explain how such hybrid architecture could provide advantages of both the instance-based and embedding-based methods. 

\subsection{DSMIL for WSI Classification}

Our key innovations are the design of a novel aggregation function $g$, and the learning of the feature extractor $f$. Specifically, we propose DSMIL that consists of a masked non-local block and a max-pooling block for feature aggregation, with input instance embeddings learned by self-supervised contrastive learning. Moreover, DSMIL combines multiscale embeddings using a pyramidal strategy, and thus ensures the local constraints of the attentions for patches in a WSI. Figure~\ref{fig:dsmil} presents an overview of our DSMIL. We now describe each component of DSMIL.

\noindent \textbf{MIL Aggregator with Masked Non-Local Operation}.
In contrast to most previous methods that either learn an instance classifier or a bag classifier, DSMIL jointly learns the instance classifier and the bag classifier as well as the bag embedding in a dual-stream architecture. 
% The evidence acquired from the instance classifier is used to produce a bag embedding for the bag classifier.

\begin{figure}
    \centering
    \includegraphics[width=0.4\textwidth]{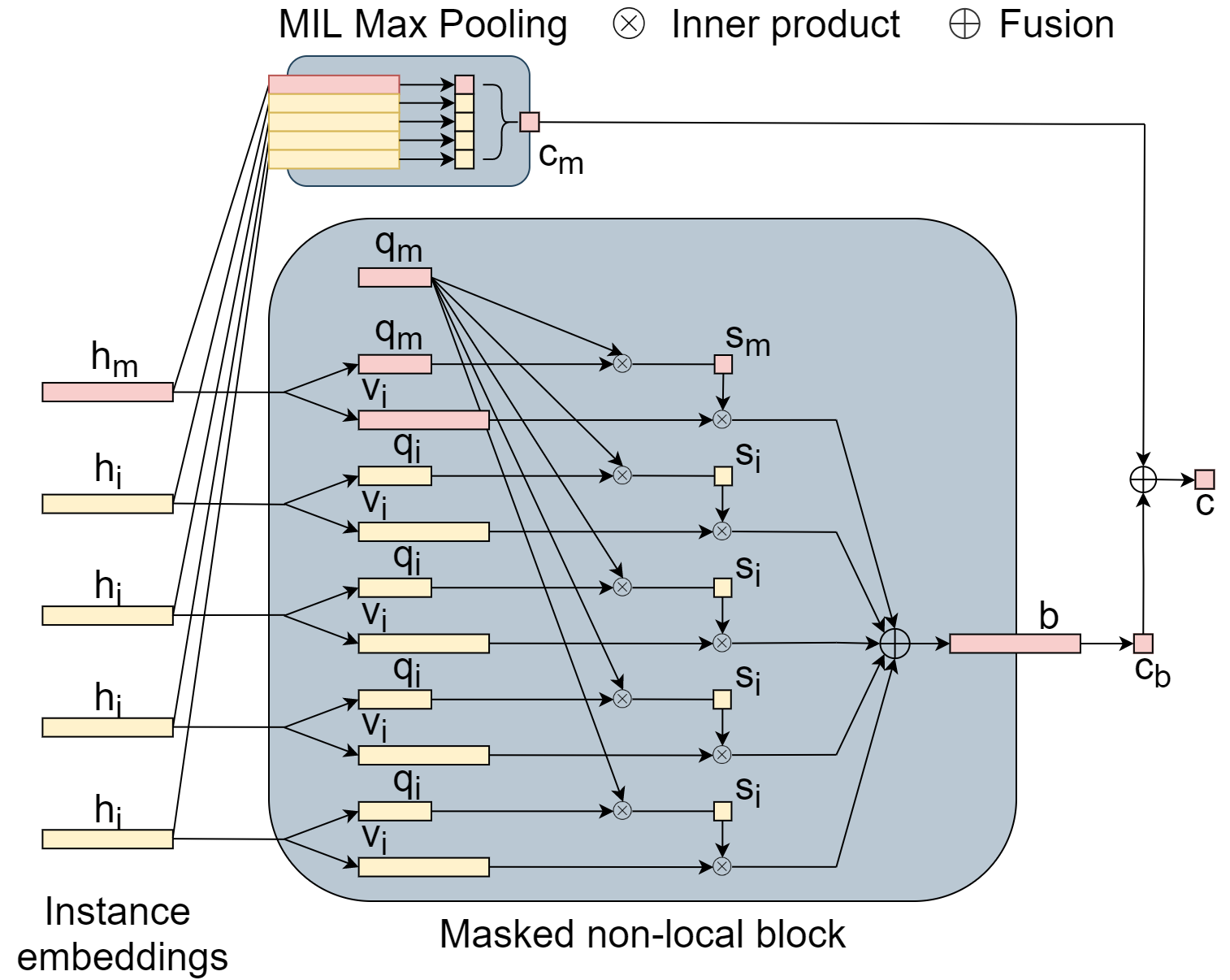}
    \caption{MIL aggregator of DSMIL. The max-pooling branch determines the critical instance by pooling the instance scores. The aggregation branch measures the distance between each instance to the critical instance and produces a bag embedding by summing the instance embeddings using the distances as weights. Scores of the two streams are averaged to produce the final score.}
    \label{fig:aggregator}\vspace{-1.5em}
\end{figure}

Let $B=\{x_1,...,x_n\}$ denote a bag of patches of a WSI. Given a feature extractor $f$, each instance $x_i$ can be projected into an embedding $\mathbf{h}_i=f(x_i)\in\mathbb{R}^{L \times 1}$. The first stream uses an instance classifier on each instance embedding, followed by max-pooling on the scores:
\begin{equation}
\small
\begin{split}
    c_m(B) {} & = g_m(f(x_i), ..., f(x_n)) \\
              & = \mathrm{max} \{ \mathbf{W}_0 \mathbf{h}_0, \dots, \mathbf{W}_{0} \mathbf{h}_{N-1}\}
\end{split}
\end{equation}
where $\mathbf{W}_0$ is a weight vector. The max-pooling stream determines the instance with the highest score (critical instance). Max-pooling is a permutation-invariant operation, thus, this stream satisfies equation \ref{eq:mil}.

The second stream aggregates the above instance embeddings into a bag embedding which is further scored by a bag classifier. We obtain the embedding $\mathbf{h}_{m}$ of the critical instance, and transform each instance embedding $\mathbf{h}_{i}$ (including $\mathbf{h}_{m}$) into two vectors, query $\mathbf{q}_{i}\in \mathbb{R}^{L \times 1}$ and information $\mathbf{v}_{i}\in \mathbb{R}^{L \times 1}$, which are given respectively by:
\begin{equation}
\small
    \mathbf{q}_{i} = \mathbf{W}_q \mathbf{h}_{i}, \quad \mathbf{v}_{i} = \mathbf{W}_v \mathbf{h}_{i}, \quad
    i=0, \dots, N-1
\end{equation}
where $\mathbf{W}_q$ and $\mathbf{W}_v$ each is a weight matrix. We then define a distance measurement $U$ between an arbitrary instance to the critical instance as:
\begin{equation}
\small
    U(\mathbf{h}_{i}, \mathbf{h}_{m}) = \frac{\exp(\langle \mathbf{q}_{i},  \mathbf{q}_{m} \rangle)}
    {\sum_{k=0}^{N-1} \mathrm{exp}(\langle \mathbf{q}_{k},  \mathbf{q}_{m} \rangle)}
    \label{eq:U}
\end{equation}
"$\langle \cdot,\cdot \rangle$" denotes the inner product of two vectors. 
The bag embedding $b$ is the weighted element-wise sum of the information vectors $\mathbf{v}_{i}$ of all instances, using the distances to the critical instance as the weights:
\begin{equation}
\small
    b = \sum_{i}^{N-1} U(\mathbf{h}_i, \mathbf{h}_m)  \mathbf{v}_{i}
\end{equation}
The bag score $c_b$ is then given by:
\begin{equation}
\small
\begin{split}
    c_b(B) {} & = g_b(f(x_i), ..., f(x_n)) \\
              & = \mathbf{W}_b \sum_{i}^{N-1} U(\mathbf{h}_i, \mathbf{h}_m) \mathbf{v}_{i} = \mathbf{W}_b \mathbf{b}
\end{split}
\end{equation}
where $\mathbf{W}_b$ is a weight vector for binary classification.
This operation is similar to self-attention~\cite{vaswani_attention_2017}. 
but differs in that the query-key matching is performed only between the critical node and the other nodes (including the critical node itself). 
Moreover, instead of matching each query with additional key vectors like self-attention, the query is matched with other queries and no key vector is learned. 

The dot product measures the similarity between two queries, resulting in larger values for instances that are more similar. Therefore, instances more similar to the critical instance will have greater attention weights. The additional layer for the information vectors $v_i$ allows contributing information to be extracted within each instance. The softmax operation in Equation~\ref{eq:U} ensures the attention weights are summed to 1 regardless of the bag size.

Since the critical instance does not depend on the order of the instances and the measurement $U$ is symmetric, this sum term so as the bag embedding $\mathbf{b}$ does not depend on the order of the instances, thus, the second stream is permutation-invariant and satisfies Equation \ref{eq:mil}. The final bag score is the average of the scores of the two streams:
\begin{equation}
\small
\begin{split}
    c(B)  & = \frac{1}{2}(g_m(f(x_i),...,f(x_n)) + g_b(f(x_i),...,f(x_n)) \\
            & = \frac{1}{2}(\mathbf{W}_0 \mathbf{h}_{m} + \mathbf{W}_b \sum_{i} U(\mathbf{h}_i, \mathbf{h}_m)  \mathbf{v}_{i}) \\
\end{split}
\end{equation}

Note that DSMIL can handle the case of multi-class MIL problems by max-pooling the instance scores and compute attention weights for each class separately. 
The result bag embedding is then a matrix $\mathrm{\mathbf{b}} \in \mathbb{R}^{L \times C}$ where $C$ is the number of classes, with each entry a weighted sum of the instance information vectors $\mathbf{v}_{i}$. 
The last fully connected layer will then have an output channel number of $C$. 

% The final score of the bag is the average of the scores of the two streams:
% \begin{equation}
% \small
%     \hat{c} = \frac{1}{2}(c_m + c_b),
% \end{equation}
The information vector $\mathbf{v}_{i}$ allows intra-instance feature selection while the distance measurement applies an inter-instance selection according to the similarity to the critical instance. 
The resulted bag embedding has a constant shape regardless of the bag size, and will be used to compute the output bag score $c_b$ at inference time.
%Note that DSMIL can use only the bag score $c_b$ as output in testing.
% information to be extracted element-wisely from each instance embedding according to the similarity to the highest-score embedding and produces a bag embedding with a constant shape regardless of the bag size.
The architecture of the aggregator is illustrated in Figure \ref{fig:aggregator}.

\noindent \textbf{Self-Supervised Contrastive Learning of WSI Features}.
Moving beyond the aggregation operation, we propose to use self-supervised contrastive learning for learning the feature extractor $f$. Specifically, we consider SimCLR from~\cite{chen_simple_2020}, a state-of-the-art self-supervised learning framework that enables robust representations to be learned without the need for manual labels. 
SimCLR deploys a contrastive learning strategy that trains the CNN to associate the sub-images from the same image in a batch of sub-images. 
The sub-images are randomly selected in a batch of images and fed into two random image augmentation branches. 
The model is trained to maximize the agreement between the sub-images that are from the same image using a contrastive loss. 
After the training converges, the feature extractor is kept and used to compute the representations of the training samples for downstream tasks. The datasets used for SimCLR consist of patches extracted from the WSIs. The patches are densely cropped without overlap and treated as individual images for SimCLR training. 
%This turns out to be appropriate since the probability to sample two adjacent patches in a mini-batch is very low. 

% Histopathological assessment involves the assessment of the sample in different microscopic scales by pathologists. 
% Region of interests are usually determined in low magnifications and each region of interests is then examined in higher magnifications. 
% Based on this observation, 
\begin{figure}
    \centering
    \includegraphics[width=0.35\textwidth]{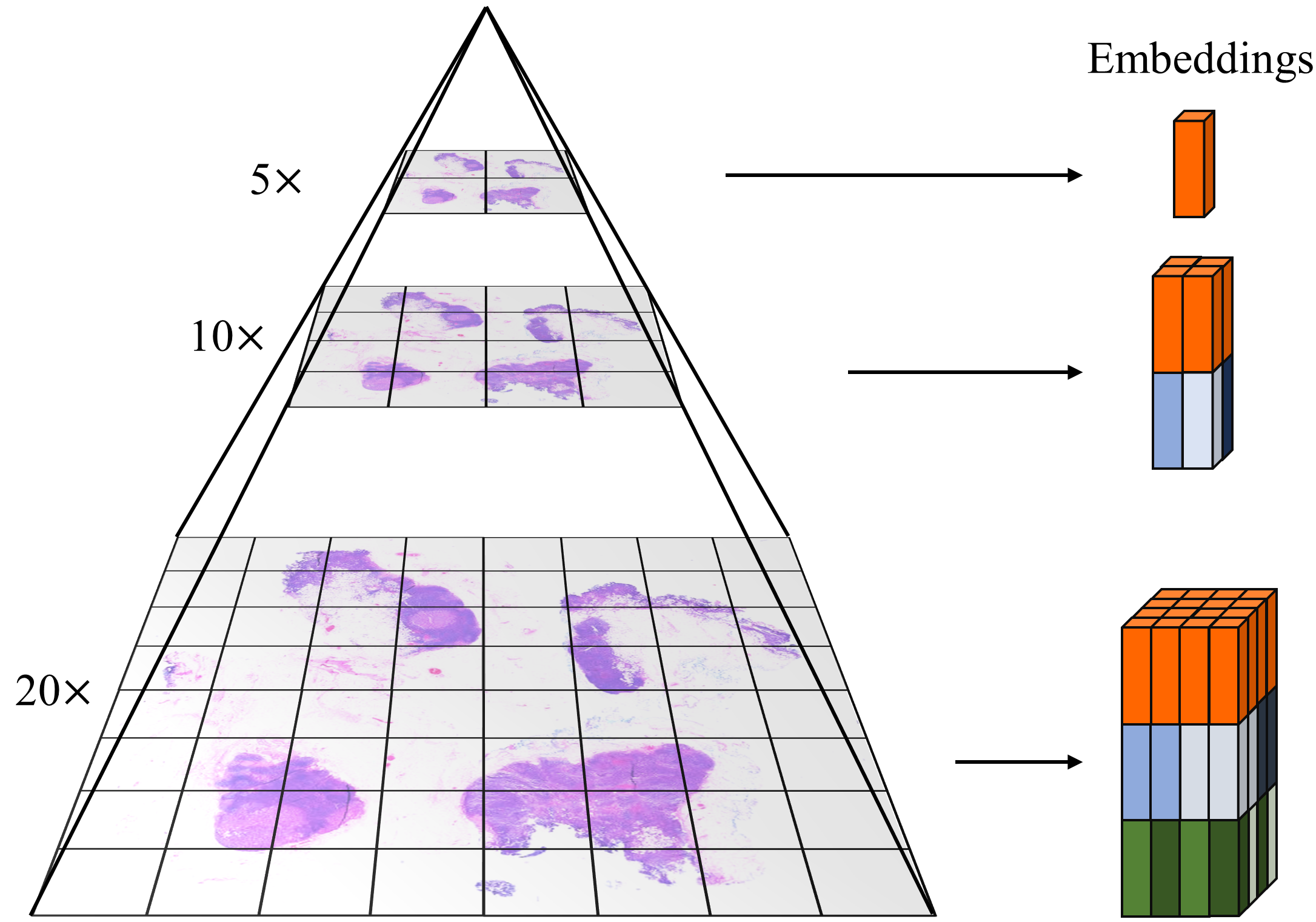}
    \caption{Pyramidal concatenation of multiscale features in WSI. Feature vector from a lower magnification patch is duplicated and concatenated to feature vectors of its higher magnification patches.}
    \label{fig:pyramid}\vspace{-1.5em}
\end{figure}

% that ensures the local constraints of the patch attentions across multiple magnifications while exposes the features from all magnifications to the network. 
% We demonstrate the effectiveness of this method using features from two magnifications, but the idea is general and can be extended to more magnifications.

\noindent \textbf{Locally-Constrained Multiscale Attention}. Finally, we make use of a pyramidal concatenation strategy to integrate features of WSIs from different magnifications.
First, For each low-magnification patch, we obtain the feature vector of this patch as well as the feature vectors of the sub-images in the higher magnification within this patch.
For example, a patch with a size of 224$\times$224 at 10$\times$ magnification will contain 4 sub-images with a size of 224$\times$224 at 20$\times$ magnification. 
For every 10$\times$ patch, we then concatenate the 10$\times$ feature vector with each of the 20$\times$ features and obtain 4 feature vectors. Figure \ref{fig:pyramid} illustrates the case of three magnifications (20$\times$-10$\times$-5$\times$).
We demonstrate the effectiveness of this method using features from two magnifications (20$\times$ and 5$\times$) in the experiment, but the idea is general and can be extended to more magnifications.

There are two major benefits of this concatenation method: 1) The first part of the resulted feature vector is the same for the 20$\times$ patches that belong to the same 5$\times$ patch. 
As a result, in DSMIL, the distance measurement results $s_i= \langle \mathbf{q}_i, \mathbf{q}_m \rangle$ for these vectors will tend to be similar and the instances will be assigned similar attention weights. 
The second part of the feature vector is specific to each 20$\times$ patch which allows the attention weights to vary among these patches. 
2) The information from different scales are preserved in the feature vectors, allowing the network to select the appropriate information $\mathbf{v}_{i}$ across different scales. 
% and the result is guaranteed to be better than or at least the same as using information solely from either scale. 
% The training procedure of the proposed model for multi-scale WSI classification is illustrated in Figure \ref{fig:pyramid}.

\section{Experiments and Results}
\label{sec:exp}
We now present our experiments and results. First, we report results on two clinical WSI datasets, Camelyon16 and TCGA lung cancer, that cover the cases of unbalanced/balanced bags and single/multiple class MIL problems. Moreover, we present an ablation study, demonstrating the effectiveness of our MIL aggregator, the contrastive feature learning, and the multiscale attention mechanism. 

\noindent \textbf{Experiment Setup and Evaluation Metrics}. We report the accuracy and area under the curve (AUC) scores of DSMIL for the task of WSI classification on both datasets. On Camelyon16, we also evaluate localization performance by reporting free response operating characteristic curves (FROC) \cite{ehteshami_bejnordi_diagnostic_2017}. To pre-process the WSIs datasets, every WSI is cropped into $224\times224$ patches without overlap to form a bag, in the magnifications of $20\times$ and $5\times$. Background patches (entropy $<5$) are discarded. Constantly better results are obtained on 20$\times$ images for both datasets and are reported for experiments using a single-scale of WSI.

\noindent \textbf{Implementation Details}. We use Adam \cite{kingma_adam_iclr} optimizer with a constant learning rate of 0.0001 to update the model weights during the training. The mini-batch size for training MIL models is 1 (bag). Patches extracted from the training sets of the WSI datasets are used to train the feature extractor using SimCLR. For SimCLR, we use Adam optimizer with an initial learning rate of 0.0001, a cosine annealing (without warm restarts) scheme for learning rate scheduling \cite{loshchilov2016sgdr}, and a min-batch size of 512. The CNN backbone used for MIL models and SimCLR is ResNet18 \cite{he_deep_2016}.

% carry out experiments on both clinical WSI datasets and standard MIL datasets. The two WSI datasets cover the cases of unbalanced/balanced bags and single/multiple class MIL problems. We report the accuracy and area under the curve (AUC) scores of DSMIL for the task of WSI classification. To preprocess the WSIs datasets, every WSI is cropped into $224\times224$ patches without overlap to form a bag, in the magnifications of $20\times$ and $5\times$. Background patches (entropy $<5$) are discarded. Constantly better results are obtained on 20$\times$ images for both datasets, so the results for 5$\times$ images are not reported for single-scale experiments.
% We use Adam \cite{kingma_adam_2017} optimizer with a constant learning rate of 0.0001 to update the model weights during the training. The mini-batch size for training MIL models is 1. 

% Patches extracted from the training sets of the WSI datasets are used to train the feature extractor using SimCLR. We use Adam optimizer with an initial learning rate of 0.0001 and a cosine annealing (without warm restarts) scheme for learning rate scheduling \cite{loshchilov_sgdr_2017}. The mini-batch size used for SimCLR is 512. 

% The CNN backbone used for MIL models and SimCLR is ResNet18 \cite{he_deep_2016}.

\subsection{Results on Camelyon16}
We first present our results on Camelyon16. We introduce the dataset and baselines, and discuss our results on both classification and localization. 

\begin{figure*}[h]
    \centering
    \includegraphics[width=0.9\textwidth]{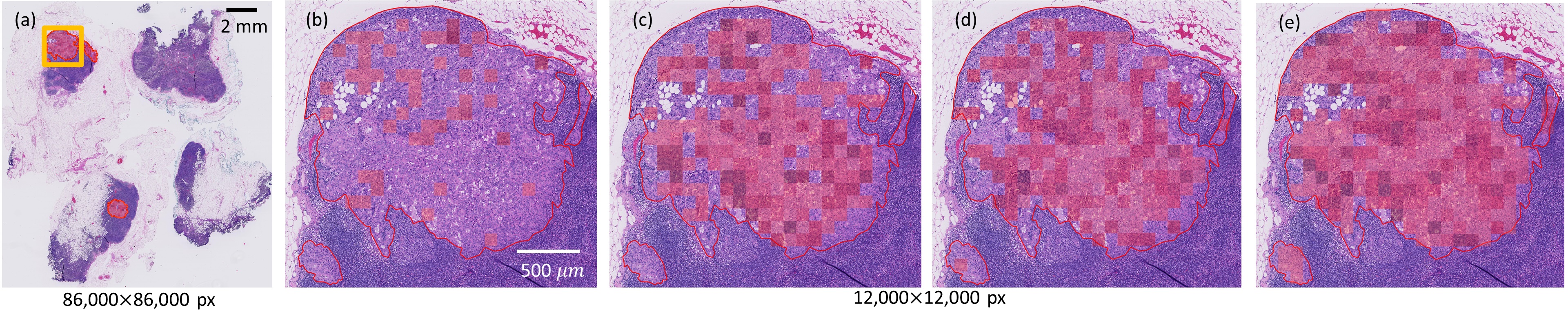}
    \caption{Tumor localization in WSI using different MIL models. \textbf{(a)} A WSI from Camelyon16 testing set. \textbf{(b)-(e)} zoomed in area in the orange box of \textbf{(a)}. \textbf{(b)} Max-pooling. \textbf{(c)} ABMIL~\cite{ilse_attention-based_2018}. \textbf{(d)} DSMIL. \textbf{(e)} DSMIL-LC \textbf{Note:} for \textbf{(b)}, classifier confidence scores are used for patch intensities; for \textbf{(c) (d)} and \textbf{(e)}, attention weights are re-scaled from min-max to [0, 1] and used for patch intensities.}
    \label{fig:localization}\vspace{-1.5em}
\end{figure*}

\begin{table}[h!]
    \centering
    \small
    \resizebox{0.48\textwidth}{!}{
    \begin{tabular}{c |c| c c| c}
    \toprule
    \multirow{2}{*}{Model} & \multirow{2}{*}{Scale} & \multicolumn{2}{c|}{Classification}             & Localization         \\
                       &                        & Accuracy             & \multicolumn{1}{c|}{AUC} & FROC                 \\
    \hline
        Mean-pooling & Single & 0.7984 & 0.7620 & 0.1162 \\
        Max-pooling & Single & 0.8295 & 0.8641 & 0.3313 \\
        MILRNN \cite{campanella_clinical-grade_2019} & Single & 0.8062 & 0.8064 & 0.3048 \\
        ABMIL \cite{ilse_attention-based_2018} & Single & 0.8450 & 0.8653 & 0.4056 \\
        DSMIL & Single & \textcolor{red}{\textbf{0.8682}} & \textcolor{red}{\textbf{0.8944}} & \textcolor{red}{\textbf{0.4296}} \\
    \hline
        Fully-supervised & Single & \textbf{0.9147} & \textbf{0.9362} & \textbf{0.5254}\\
    \hline
        MS-MILRNN \cite{campanella_clinical-grade_2019} & Multiple & 0.8140 & 0.8371 & 0.2791 \\ 
        MS-ABMIL \cite{hashimoto_multi-scale_2020} & Multiple & 0.8760 & 0.8872 & 0.4191 \\
        DSMIL-LC & Multiple & \textcolor{red}{\textbf{0.8992}} & \textcolor{red}{\textbf{0.9165}} & \textcolor{red}{\textbf{0.4371}} \\ 
    \bottomrule
    \end{tabular}
    }\vspace{0.1em}
    \caption{Results on Camelyon16 dataset. DSMIL/DSMIL-LC denote our model with/without the proposed multiscale attention mechanism. Instance embeddings are produced by the feature extractor trained using SimCLR for all MIL models.}
    \label{tab:c16_results}\vspace{-1em}
\end{table}

\noindent \textbf{Dataset}. Camelyon16 is a public dataset proposed for metastasis detection in breast cancer \cite{ehteshami_bejnordi_diagnostic_2017}. The dataset consists of 271 training images and 129 testing images, which yield roughly 3.2 million patches at 20$\times$ magnification and 0.25 million patches at 5$\times$ magnification with on average about 8,000 and 625 patches per bag. Tumor regions are fully labeled with pixel-level annotations on each slide. We ignore the pixel-level annotations in the training and consider only slide-level labels (\ie the slide is considered positive if it contains any annotated tumor regions). The resulted bags contain mixtures of tumor and healthy patches for positive bags and all healthy patches for negative bags.

The positive bags in this dataset are highly unbalanced. Only a small portion of regions in a positive slide contains tumor (roughly $<$10\% of the total tissue area per slide) which leads to a large portion of negative patches in a positive bag.
This makes it hard for good representations to be directly learned in most MIL models \cite{lu_semi-supervised_2019, dehaene_self-supervision_2020}.
% Though there exist published studies that use region-level labels instead of pixel-level labels \cite{shen_deep_2019,xu_camel_2019} with weakly-supervised learning on this dataset, studies that only consider the slide-level labels are few. 
We show that our method relying on only the slide-level labels can overcome this difficulty and achieves performance comparable to fully-supervised methods that use the pixel-level labels.

\noindent \textbf{Baselines}. We evaluate and compare DSMIL to a strong set of baselines, including (1) deep models using traditional MIL pooling operators such as max-pooling and mean-pooling and (2) recent deep MIL models~\cite{hashimoto_multi-scale_2020,campanella_clinical-grade_2019,ilse_attention-based_2018}, on the tasks of WSI classification and tumor localization. Moreover, we obtain an upper-bound fully-supervised model by making use of the pixel-level annotations, where a patch is labeled positive if it falls within a tumor region and the score of a WSI is then obtained by averaging the scores of all its patches in testing. Results on the classification task can demonstrate the efficacy of our model in terms of producing good bag embeddings, while results on the localization task can demonstrate the capability of DSMIL to delineate positive instances in positive bags. %WSIs usually lead to bags with a few hundred to a couple of thousands of patches, which poses extra challenges for MIL models to learn non-trivial aggregation functions as well as good decision boundaries for the instances. 

\noindent \textbf{Classification Results}. The classification results are summarized in Table \ref{tab:c16_results}.
Features are learned using self-supervised contrastive learning on the $20\times$ patches under the same settings. The contribution of using self-supervised contrastive learning will be presented in the ablation study. The results suggest that, though both better than traditional pooling operators, DSMIL achieves better aggregation than ABMIL which implements no additional regularization on the learned attentions, with about 2.6\% improvements in classification on the single scale setting. 
The recurrent neural network-based model without considering the permutation-invariant characteristics does not outperform the traditional pooling operators. With the multiscale attention mechanism integrated, DSMIL achieves improved results matching the performance of the fully-supervised method, with a classification accuracy gap smaller than 2\%.

\noindent \textbf{Localization Results}. Pixel-level annotations are available for Camelyon16 which allow us to test the localization ability of our method. 
The localization performance indicates the MIL model's capability to delineate positive instances.
% We use FROC to evaluate the detection localization \cite{ehteshami_bejnordi_diagnostic_2017} of both MIL-based methods and the fully-supervised upper bound. The results are summarized in Table \ref{tab:c16_results}. 
The reported FROC score is defined as the average sensitivity at 6 predefined false positive rates: 1/4, 1/2, 1, 2, 4, and 8 FPs per WSI. 
The result shows that DSMIL, where the attention scores are explicitly computed using a trainable distance measurement, better delineates the positive patches with at least 6\% relative improvement compared to ABMIL in detection localization. Detection maps of representative samples from the testing set are illustrated in Figure \ref{fig:localization}. 

\subsection{Results on TCGA Lung Cancer dataset}
We further present our results on The Cancer Genome Atlas (TCGA) lung cancer dataset. We again introduce the dataset and discuss our results.

\begin{table}[h!]
    \centering
    \footnotesize
    \begin{tabular}{c | c | c c }
    \toprule
        \multicolumn{4}{c}{\textbf{SimCLR features}} \\
    \hline 
        Model & Scale & Accuracy & AUC \\
    \hline
        Mean-pooling & Single & 0.8857 & 0.9369 \\
        Max-pooling & Single & 0.8088 & 0.9014 \\
        MIL-RNN \cite{campanella_clinical-grade_2019} & Single & 0.8619 & 0.9107 \\
        ABMIL \cite{ilse_attention-based_2018} & Single & 0.9000 & 0.9488 \\
        DSMIL & Single & \textcolor{red}{\textbf{0.9190}} & \textcolor{red}{\textbf{0.9633}} \\
    \hline
        MS-MIL-RNN \cite{campanella_clinical-grade_2019} & Multiple & 0.8905 & 0.9213 \\
        MS-ABMIL \cite{hashimoto_multi-scale_2020} & Multiple & 0.9000 & 0.9551 \\
        DSMIL-LC & Multiple & \textcolor{red}{\textbf{0.9286}} & \textcolor{red}{\textbf{0.9583}} \\ 
    \toprule
        \multicolumn{4}{c}{\textbf{Patch-based features}} \\
    \hline 
        Model & Scale & Accuracy & AUC \\
    \hline
        Patch-based w/o MIL & Single & 0.8857 & 0.9506 \\
    \hline
        Mean-pooling & Single & 0.9096 & 0.9625 \\
        Max-pooling & Single & 0.8286 & 0.8958 \\
        MIL-RNN \cite{campanella_clinical-grade_2019} & Single & 0.9048 & 0.9636 \\
        ABMIL \cite{ilse_attention-based_2018} & Single & 0.9381 & 0.9765 \\
        DSMIL & Single & \textcolor{red}{\textbf{0.9476}} & \textcolor{red}{\textbf{0.9809}} \\
    \hline
        MS-MIL-RNN \cite{campanella_clinical-grade_2019} & Multiple & 0.9096 & 0.9561 \\
        MS-ABMIL \cite{hashimoto_multi-scale_2020} & Multiple & 0.9381 & 0.9792 \\
        DSMIL-LC & Multiple & \textcolor{red}{\textbf{0.9571}} & \textcolor{red}{\textbf{0.9815}}\\
    \bottomrule 
    \end{tabular} \vspace{0.1em}
    \caption{Results on TCGA lung cancer dataset. Instance embeddings are produced by the feature extractor trained using SimCLR and patch-based method without considering MIL.}
    \label{tab:tcga_results}\vspace{-1em}
\end{table}

\noindent \textbf{Dataset}. The WSIs include two sub-types of lung cancer, Lung Adenocarcinoma and Lung Squamous Cell Carcinoma, with in a total of 1054 diagnostic digital slides that can be downloaded from National Cancer Institute Data Portal. We randomly split the WSIs into 840 training slides and 210 testing slides (4 low-quality corrupted slides are discarded). The dataset yields 5.2 million patches at 20$\times$ magnification and 0.36 million patches at 5$\times$ magnification with in average about 5000 and 350 patches per bag. Only slide-level labels are available for this dataset.

The resulted bags contain mixtures of either type of tumor and healthy patches for positive bags, and all healthy patches for negative bags.
Tumor slides in this dataset contain large portions of tumor regions ($>$80\% per slide), leading to a large portion of positive patches in positive bags. Thus, training a classifier using a patch-based method without considering MIL already has reasonable results (\ie treating the patches in a WSI as if they all have the same label as the whole WSI in training, and averaging the scores of the patches in a WSI in testing). We show that significantly improved results can be obtained by considering MIL.

\noindent \textbf{Classification Results}. We compare both the features learned by SimCLR and by the patch-based method without considering MIL for this dataset. By contrast, the patch-based method does not converge for Camelyon16 due to the large number of negative patches in positive bags, so the patch-based features results are not included for Camelyon16. The results are summarized in Table \ref{tab:tcga_results} which suggests similar conclusions as Camelyon16 dataset.

\subsection{Ablation Study}
We now delineate the contributions of our model via ablation studies of the three major components of our model: DSMIL aggregator, self-supervised contrastive learning for the instance features, and the multiscale attention mechanism. We keep our DSMIL aggregator and compare features learned by different methods as well as different multiscale feature fusion methods for WSI.
While the performance of our DSMIL aggregator has been demonstrated on two WSI datasets in the previous section, we further carry out extensive benchmark experiments for our MIL aggregator on several classical MIL datasets in the ablation study.

\begin{table}[th!]
    \centering
    \footnotesize
    \begin{tabular}{c | c c | c c}
    \toprule
        Dataset & \multicolumn{2}{c|}{\textbf{Camelyon16}} & \multicolumn{2}{c}{\textbf{TCGA}} \\
    \hline
        Features & Accuracy & AUC &  Accuracy & AUC \\
    \hline
        ImageNet  & 0.6202 & 0.5408 & 0.7095 & 0.7260 \\ 
        Max-pooling  & 0.7099 & 0.7153 & 0.7714 & 0.8212 \\
        Patch-based  & 0.6977 & 0.5434 & \textcolor{red}{\textbf{0.9476}} & \textcolor{red}{\textbf{0.9809}} \\
        Contrastive  & \textcolor{red}{\textbf{0.8682}} & \textcolor{red}{\textbf{0.8944}} & \textcolor{blue}{\textbf{0.9190}} & \textcolor{blue}{\textbf{0.9633}} \\
    % \toprule
    %     \multicolumn{3}{c}{\textbf{TCGA lung cancer}} \\
    % \hline
    %     Features & Accuracy & AUC \\
    % \hline
    %     ImageNet  & 0.7095 & 0.7260 \\ 
    %     Max-pooling & 0.7714 & 0.8212 \\
    %     Patch-based  & \textcolor{red}{0.9476} & \textcolor{red}{0.9809} \\
    %     Contrastive learning & \textcolor{blue}{0.9190} & \textcolor{blue}{0.9633} \\
    \bottomrule
    \end{tabular} \vspace{0.1em}
    \caption{Comparison of features learned by different methods for a fixed MIL aggregator.}
    \label{tab:features}\vspace{-1em}
\end{table}

\begin{table}[th!]
    \centering
    \footnotesize
    \begin{tabular}{c | c c}
    \toprule
        Method & Accuracy & AUC \\
    \hline
        Single scale (20$\times$) & 0.8682 & 0.8944 \\ 
        Concatenation (5$\times$ + 20$\times$) & 0.8682 & 0.8846 \\
        Max Pooling (5$\times$ + 20$\times$)  & 0.8604 & 0.8731 \\
        Mix (5$\times$ + 20$\times$)  & 0.8837 & 0.9097 \\ \hline
        Ours (5$\times$ + 20$\times$) & \textcolor{red}{\textbf{0.8992}} & \textcolor{red}{\textbf{0.9165}} \\
        Ours (1.25$\times$ + 5$\times$ + 20$\times$) & 0.8760 & 0.9034 \\
    \bottomrule
    \end{tabular}\vspace{0.1em}
    \caption{Comparison of different multiscale WSI feature integration methods. Multiscale approaches from other studies are used on our MIL aggregator with fixed instance embeddings learned by self-supervised contrastive learning on 20$\times$ and 5$\times$ WSI patches.}\vspace{-1em}
    \label{tab:multiscale}
\end{table}

\noindent \textbf{Effects of Contrastive Learning}.
% The feature extractor for MIL often needs to be pre-trained and used to compute fixed features before training the MIL aggregator for WSI, due to the prohibitive memory requirement~\cite{campanella_clinical-grade_2019, lu_semi-supervised_2019, akbar_cluster-based_2018}. 
We compare the features learned by self-supervised contrastive learning to several baselines. 1) Use the feature extractor trained by max-pooling operator~\cite{campanella_clinical-grade_2019}. The end-to-end training using max-pooling can be done in a for-loop where the maximum-score instance is found dynamically and used to update the model weights without the need for large memory. 2) Use the feature extractor trained by the patch-based method without considering MIL (\ie treating the patches in a WSI as if they all have the same label as the WSI in training, and averaging the scores of the patches in a WSI in testing). 3) Use the feature extractor pre-trained on ImageNet dataset~\cite{deng_imagenet_2009}. 

\begin{table*}[t!]
    \centering
    \footnotesize
    \begin{tabular}{c | c c c c c}
    \toprule
         Methods & MUSK1 & MUSK2 & FOX & TIGER & ELEPHANT \\
         \hline
         mi-Net & 0.889 $\pm$ 0.039 & 0.858 $\pm$ 0.049 & 0.613 $\pm$ 0.035 & 0.824 $\pm$ 0.034 &  0.858 $\pm$ 0.037 \\
         MI-Net & 0.887 $\pm$ 0.041 & 0.859 $\pm$ 0.046 & 0.622 $\pm$ 0.038 & 0.830 $\pm$ 0.032 &  0.862 $\pm$ 0.034 \\
         MI-Net with DS & 0.894 $\pm$ 0.042 & 0.874 $\pm$ 0.043 & 0.630 $\pm$ 0.037 & 0.845 $\pm$ 0.039 &  0.872 $\pm$ 0.032 \\
         MI-Net with RC & 0.898 $\pm$ 0.043 & 0.873 $\pm$ 0.044 & 0.619 $\pm$ 0.047 & 0.836 $\pm$ 0.037 &  0.857 $\pm$ 0.040 \\
         ABMIL & 0.892 $\pm$ 0.040 & 0.858 $\pm$ 0.048 & 0.615 $\pm$ 0.043 & 0.839 $\pm$ 0.022 &  0.868 $\pm$ 0.022 \\
         ABMIL-Gated & 0.900 $\pm$ 0.050 & 0.863 $\pm$ 0.042 & 0.603 $\pm$ 0.029 & 0.845 $\pm$ 0.018 &  0.857 $\pm$ 0.027 \\
         GNN-MIL & 0.917 $\pm$ 0.048 & 0.892 $\pm$ 0.011 & 0.679 $\pm$ 0.007 & 0.876 $\pm$ 0.015 &  0.903 $\pm$ 0.010 \\
         DP-MINN & 0.907 $\pm$ 0.036 & 0.926 $\pm$ 0.043 & 0.655 $\pm$ 0.052 & \textcolor{red}{\textbf{0.897 $\pm$ 0.028}} &  0.894 $\pm$ 0.030 \\
         \hline
         NLMIL & 0.921 $\pm$ 0.017 & 0.910 $\pm$ 0.009 & 0.703 $\pm$ 0.035 & 0.857 $\pm$ 0.013 &  0.876 $\pm$ 0.011 \\
         ANLMIL & 0.912 $\pm$ 0.009 & 0.822 $\pm$ 0.084 & 0.643 $\pm$ 0.012 & 0.733 $\pm$ 0.068 &  0.883 $\pm$ 0.014 \\
         \hline 
         DSMIL & \textcolor{red}{\textbf{0.932 $\pm$ 0.023}} & \textcolor{red}{\textbf{0.930 $\pm$ 0.020}} & \textcolor{red}{\textbf{0.729 $\pm$ 0.018}} & {0.869 $\pm$ 0.008} &  \textcolor{red}{\textbf{0.925 $\pm$ 0.007}} \\
         \bottomrule
    \end{tabular} \vspace{0.1em}
    \caption{Performance comparison on classical MIL dataset. Experiments were run 5 times each with a 10-fold cross-validation. The mean and standard deviation of the classification accuracy is reported (mean $\pm$ std). mi-Net\cite{wang_revisiting_2018}, MI-Net \cite{wang_revisiting_2018}, MI-Net with DS \cite{wang_revisiting_2018}, MI-Net with RC \cite{wang_revisiting_2018}, ABMILP \cite{ilse_attention-based_2018}, ABMILP-Gated \cite{ilse_attention-based_2018}, GNN-MIL \cite{tu_multiple_2019}, DP-MINN \cite{yan_deep_2018}. NLMIL and ANLMIL use the non-local blocks from~\cite{wang_non-local_2018} and~\cite{zhu_asymmetric_2019}. Previous benchmark results are taken from \cite{ilse_attention-based_2018, tu_multiple_2019, yan_deep_2018} and the same training setting as \cite{ilse_attention-based_2018} is used.}
    \label{tab:classic} \vspace{-1em}
\end{table*}

The results are shown in Table \ref{tab:features}. For unbalanced bags (\eg, Camelyon16 dataset), self-supervised contrastive learning leads to significantly better performance with at least 16\% higher classification accuracy, even compared to the features obtained by end-to-end training of max-pooling. For balanced bags (\eg, TCGA lung cancer dataset), features learned by self-supervised contrastive learning are comparable to those of the patch-based method, yet are still significantly better ($>14\%$ higher accuracy) than end-to-end training of max-pooling. Note that for unbalanced bags, the patch-based method does not lead to good features due to large amounts of negative samples in positive bags.
Moreover, we further observe that using max-pooling on contrastive learning features also significantly outperforms the end-to-end training of max-pooling by about $10\%$. 
The results suggest that self-supervised contrastive learning is a feasible way to obtain good representations for MIL regardless of the distribution of negative and positive instances in the bags, and it also alleviates the memory requirement issue caused by large bag size.

\noindent \textbf{Effects of Multiscale Attention}.
We further compare our multiscale attention mechanism to several other methods that consider multiscale WSI features, including 1) Concatenate the bag embeddings of the MIL model trained on each magnification before the fully-connected layer. 
2) Use max-pooling on the predictions of the MIL model trained on each magnification \cite{campanella_clinical-grade_2019}.
3) Mix the instances from different scales in a bag and feed the bag to the MIL model~\cite{hashimoto_multi-scale_2020}. 

Table \ref{tab:multiscale} presents the results on Camelyon16 dataset. Our multiscale attention outperforms the single scale approach by 3\% and other multiscale approaches by at least 1.5\%, suggesting that considering multiscale features could lead to better detection accuracy for WSI and structured multiscale features can further improve the results. Yet using two levels (5$\times$+20$\times$) produces better results than using all three levels (1.25$\times$+5$\times$+20$\times$) with +1.6\% in accuracy and +1.3\% in AUC. We conjecture that sometimes information from a coarser scale (\eg 1.25$\times$) might not be as effective as a finer one (\eg 20$\times$), and the resulted vectors could become less discriminate. Thus, an attention mechanism along the magnification level might be needed to re-weight the features from different scales before fusion.

\noindent \textbf{DSMIL Aggregator on Other MIL Tasks}.
Finally, We benchmark our dual-stream MIL aggregator on classical MIL benchmark datasets.
These datasets consist of extracted feature vectors of the instances and do not require a feature extractor to be learned.
The first two datasets (MUSK1, MUSK2) are used to predict drug effects based on the molecule conformations. 
A molecule can have different conformations and only some of them may be effective conformations \cite{dietterich_solving_1997}.
Each bag contains multiple conformations of the same molecule, and the bag is labeled positive if at least one conformation is effective, negative otherwise. 
The other three datasets, ELEPHANT, FOX, and TIGER, consists of feature vectors extracted from images. 
Each bag includes a group of segments of an image and the bag is labeled as positive if at least one segment contains the animal of interest, negative if there is no such animal presented. 

Since the feature vectors (instance embeddings) are already given, the experiment involves directly feeding the feature vectors to DSMIL aggregator. To test our MIL aggregator against other recent non-local architectures on MIL problem, we replace the proposed DSMIL aggregator with the non-local blocks in NL~\cite{wang_non-local_2018} (NLMIL) and ANL~\cite{zhu_asymmetric_2019} (ANLMIL) and also evaluate their results across the 5 MIL datasets \ref{tab:classic}. Experiments are run 5 times each with a 10-fold cross-validation. 
% The mean and standard deviation of the classification accuracy is reported in Table \ref{tab:classic}. Experiments are run 5 times each with a 10-fold cross-validation. 
The benchmark results show that our dual-stream MIL aggregator outperforms the previous best MIL models as well as other non-local operations such as NL and ANL by an average of 3\% on general MIL problems.

\section{Conclusion and Future Work}
% yin: you should try to end strong here. Emphasis key contribution / results / impact to the field.
In this paper, we present a new MIL-based approach for weakly supervised WSI classification. Our method has demonstrated considerable improvement over previous methods on representative WSI datasets.
Our key technical innovation is a novel MIL aggregator that outperforms recent MIL models on both MIL benchmark dataset and representative WSI datasets. We also propose to make use of self-supervised contrastive learning in MIL models and to incorporate multiscale features. Our method further integrates the proposed aggregator, contrastive learning, and multiscale features into a MIL model for WSI classification.
By casting tumor detection in WSI as a MIL problem, our solution has the potential for real-world clinical applications where large amount of unannotated slides are available. We believe our work provides a solid step forward for both MIL and computational histopathology.

Future research includes designing self-supervised learning strategies that adapt to the characteristics of histopathological data. 
%The current self-supervised learning framework used in our study might be sub-optimal for histopathological images.
Moreover, mechanisms that model the spatial relations can be integrated to capture macroscale features in WSI that are spatially structured and could potentially lead to further improvement. \medskip

\noindent \textbf{Acknowledgment}: The work was supported by NIH P41-GM135019, the Semiconductor Research Corporation (SRC), and the Morgridge Institute for Research. YL also acknowledges the support by the UW VCRGE with funding from WARF. 

% The contribution of each component in our design are demonstrated by ablation studies.
% The proposed DSMIL aggregator leads to better performance on both WSI classification and general MIL problems. 
% The use of self-supervised contrastive learning alleviates the issue caused by unbalanced bags and large bags, allowing good representations to be learned for downstream MIL aggregation.
% The pyramidal multiscale feature fusion method for WSI further improves the classification and localization performance.
% Our weakly supervised approach does not require annotated WSIs for training while achieves WSI classification and tumor detection with performance matching the fully-supervised method.
% Weakly labeled WSIs are easy to obtain on large scale from everyday clinics while fully-annotated WSIs require costly expert labor. 
% Thus, our work is compatible with the standard clinical flow and could potentially generalize better than fully-supervised methods which rely on limited annotated WSIs~\cite{campanella_clinical-grade_2019}.
% % The self-supervised learning framework used in our study is a solution for general images.
% Future research includes designing self-supervised learning strategies that adapt to the characteristics of histopathological data. 
% %The current self-supervised learning framework used in our study might be sub-optimal for histopathological images.
% Moreover, mechanisms that model the spatial relations can be integrated to capture macroscale features in WSI that are spatially structured and could potentially lead to further improvement.

{\small
\bibliographystyle{ieee_fullname}
\bibliography{egbib, references}
}

\end{document}